\documentclass{article}

\usepackage[accepted]{icml2020}

\usepackage{colortbl}
\usepackage[utf8]{inputenc} 
\usepackage[T1]{fontenc}    
\usepackage{hyperref}       
\usepackage{url}            
\usepackage{booktabs}       
\usepackage{amsfonts}       
\usepackage{nicefrac}       
\usepackage{microtype}      
\usepackage[english]{babel}
\usepackage{graphicx}
\usepackage{appendix}
\usepackage{hyperref}
\usepackage{amsmath}
\usepackage{dsfont}
\usepackage{lineno}
\usepackage{url}
\usepackage{lipsum}
\usepackage{mathtools}
\usepackage{cuted}
\usepackage{graphics}
\usepackage{amsmath}
\usepackage{lipsum}

\renewcommand\footnotemark{}

\title{\LARGE \bf INSTA-YOLO: Real-Time Instance Segmentation based on YOLO}

\begin{document}

\twocolumn[
\icmltitle{INSTA-YOLO: Real-Time Instance Segmentation}




\begin{icmlauthorlist}
\icmlauthor{Eslam Mohamed}{valeo,cairo}
\icmlauthor{Abdelrahman Shaker}{valeo}
\icmlauthor{Ahmad El-Sallab}{valeo}
\icmlauthor{Mayada Hadhoud}{cairo}
\end{icmlauthorlist}

\icmlaffiliation{valeo}{Valeo, Egypt}
\icmlaffiliation{valeo}{Valeo, Egypt}
\icmlaffiliation{cairo}{Faculty of Engineering, Cairo University}

\icmlcorrespondingauthor{Eslam Mohamed}{eslam.mohamed-abdelrahman@valeo.com}
\icmlcorrespondingauthor{Abdelrahman Shaker}{abdelrahman.shaker@valeo.com}
\icmlcorrespondingauthor{Ahmad El-Sallab}{ahmad.el-sallab@valeo.com}
\icmlcorrespondingauthor{Mayada Hadhoud}{mayada.hadhoud@eng.cu.edu.eg}

\icmlkeywords{Machine Learning, ICML}

\vskip 0.3in
]

\printAffiliationsAndNotice{}


\begin{abstract}

Instance segmentation has gained recently huge attention in various computer vision applications. It aims at providing different IDs to different objects of the scene, even if they belong to the same class. Instance segmentation is usually performed as a two-stage pipeline. First, an object is detected, then semantic segmentation within the detected box area is performed which involves costly up-sampling. In this paper, we propose Insta-YOLO, a novel one-stage end-to-end deep learning model for real-time instance segmentation. Instead of pixel-wise prediction, our model predicts instances as object contours represented by 2D points in Cartesian space. We evaluate our model on three datasets, namely, Carvana, Cityscapes and Airbus. We compare our results to the state-of-the-art models for instance segmentation. The results show our model achieves competitive accuracy in terms of mAP at twice the speed on GTX-1080 GPU.

\end{abstract}


\section{Introduction}

Environment perception pipeline is required to isolate objects of interest from background ones. While multi-object detection produces a bounding box around the objects of interest, a tight contour is required especially with occluding objects. This is obtained through semantic segmentation which is incapable of distinguishing occluding object instances, and is computationally expensive due to pixel-wise operations. On the contrary, instance segmentation can produce a tight object mask, with differentiation between objects instances, without classifying the whole scene pixels.

The standard method in literature for instance segmentation is to build a two-stage pipeline \cite{he2017mask}. First, objects are detected and highlighted using bounding boxes. Then, a semantic segmentation model processes the areas of the detected boxes to produce objects masks. This approach suffers from various drawbacks: (1) Computational complexity: due to costly upsampling (2) Slow processing: since the two-stages have to run sequentially (3) Does not fit for tasks that involve oriented boxes, such as remote sensing applications \cite{tang2017arbitrary}, or Bird-eye-View LiDAR \cite{ali2018yolo3d}.

In this paper, We propose Insta-YOLO, a novel one-shot end-to-end model, inspired by YOLO one-shot object detector \cite{redmon2018yolov3}. Our model predicts instance masks represented by contour points in Cartesian space. Our method does not perform costly pixel-wise prediction or up-sampling. Hence, the algorithm is efficient in terms of number of parameters and fast in terms of FPS. Moreover, it provides generic solution for oriented objects as well. In summary, the contribution of this work includes:

\begin{itemize}
    \item Construction of a novel CNN architecture for instance segmentation. Our model  runs in real-time at double speed of the state-of-the-art algorithms with competitive accuracy.
    \item Implementation of a novel loss function that improves localization of the instances.
    \item Demonstration of how our algorithms generalizes to tasks with oriented objects.
\end{itemize}

The rest of the paper is organized as follows. First, we discuss the related work, followed by the model architecture details. Then we present the details of model training and experimental setup and discuss the results. Finally, we conclude with the main findings.


\section{Related Work}
\label{sec-related-work}

In this section we review the recent models for object detection, instance segmentation, and polygon regression techniques.

\subsection{Object Detection}
\label{sec-object-detection}

The goal of object detectors is to predict the location of any object of interest with a recognized label. Object detectors can generally be split into two main categories, two-stage, and one-stage detectors. The two-stage detectors consist of a region proposal network in the first stage. The features are extracted from each candidate box for bounding-box regression and recognition tasks in the second stage. Two-stage object detectors usually achieve higher localization and recognition accuracy, but they are not feasible for real-time applications. On the other hand, one-stage detectors localize the bounding boxes and determine their labels in an end-to-end way. Hence, they need less inference time and can be used for real-time applications.

YOLO \cite{redmon2016you}, SSD \cite{liu2016ssd} and RetinaNet  \cite{lin2017focal} are the most common one-stage object detectors. YOLO divides the image into S x S grid cells, which are responsible for detecting N objects whose centers fall within. Each grid cell predicts N boxes, every box is represented using x,y,w,h for the localization and C class probabilities. The backbone has 24 convolutional layers, followed by two fully connected layers that are responsible for object localization and recognition. Experimental results show that YOLO was not good enough to localize the objects precisely. It achieved mAP of 63.4\% with 45 FPS on PASCAL VOC dataset. Based on that, the authors proposed YOLOv2 \cite{redmon2017yolo9000} and YOLOv3 \cite{redmon2018yolov3} with a series of improvement in design choices, including different backbones, more bounding boxes, and different scales to achieve state-of-the-art accuracy. On a Pascal Titan X it processes images at 30 FPS and has a mAP of 57.9. 
Faster R-CNN  \cite{ren2015faster} is the most known two-stage object detector. It was proposed to improve the region-based CNN in R-CNN \cite{girshick2014rich} and Fast R-CNN \cite{girshick2015fast}. Instead of using selective search to generate region proposals, which need running time equivalent to the detection network, it uses a fully convolutional region proposal network to predict the candidate boxes efficiently. Faster R-CNN achieved mAP of 73.2\% with 7 FPS on PASCAL VOC dataset.

\subsection{Instance Segmentation}
\label{sec-insta-seg}

Instance segmentation requires predicting the instances of the objects and their binary segmentation mask. It performs object detection and semantic segmentation simultaneously. Existing methods in the literature are often divided into two groups, two-stage, and one-stage instance segmentation.

Two-stage instance segmentation methods usually deal with the problem as a detection task followed by segmentation, they detect the bounding boxes of the objects in the first stage, and a binary segmentation is performed for each bounding box in the second stage. Mask R-CNN \cite{he2017mask}, which is an extension for the Faster R-CNN, adds an additional branch that computes the mask to segment the objects and replaces RoI-Pooling with RoI-Align to improve the accuracy. Hence, the Mask R-CNN loss function combines the losses of the three branches; bounding box, recognized class, and the segmented mask. Mask Scoring R-CNN \cite{huang2019mask} adds a mask-IoU branch to learn the quality of the predicted masks in order to improve the performance of the instance segmentation by producing more precise mask predictions. The mAP is improved from 37.1\% to 38.3\% compared to the Mask R-CNN on the COCO dataset. PANet  \cite{liu2018path} is built upon the Mask R-CNN and FPN networks. It enhanced the extracted features in the lower layers by adding a bottom-up pathway augmentation to the FPN network in addition to proposing adaptive feature pooling to link the features grids and all feature levels. PANet achieved mAP of 40.0\% on the COCO dataset, which is higher than the Mask R-CNN using the same backbone by 2.9\% mAP. Although the two-stage methods can achieve state-of-the-art performance, they are usually quite slow and can not be used for real-time applications. Using one TITAN GPU, Mask R-CNN runs at 8.6 fps, and PANet runs at 4.7 fps. Real-time instance segmentation usually requires running above 30 fps.

One-stage methods usually perform detection and segmentation directly. InstanceFCN \cite{dai2016instance} uses FCN to produce several instance-sensitive score maps that have information for the relative location, then object instances proposals are generated by using an assembling module. YOLACT \cite{bolya2019yolact}, which is one of the first attempts for real-time instance segmentation, consists of feature backbone followed by two parallel branches. The first branch generates multiple prototype masks, whereas the second branch computes mask coefficients for each object instance. After that, the prototypes and their corresponding mask coefficients are combined linearly, followed by cropping and threshold operations to generate the final object instances. YOLACT achieved mAP of 29.8\% on the COCO dataset at 33.5 fps using Titan Xp GPU. YOLACT++  \cite{bolya2019yolact++} is an extension for YOLACT with several performance improvements while keeping it real-time. Authors utilized the same idea of the Mask Scoring R-CNN and added a fast mask re-scoring branch to assign scores to the predicted masks according to the IoU of the mask with ground-truth. Also, the 3x3 convolutions in specific layers are replaced with 3x3 deformable convolutions in the backbone network. Finally, they optimized the prediction head by using multi-scale anchors with different aspect ratios for each FPN level.
YOLACT++ achieved 34.1\% mAP (more than YOLACT by 4.3\%) on the COCO dataset at 33.5 fps using Titan Xp GPU. However, the deformable convolution makes the network slower, as well as the upsampling blocks in YOLACT networks.

TensorMask \cite{chen2019tensormask} explored the dense sliding-window instance segmentation paradigm by utilizing structured 4D tensors over the spatial domain. Also, tensor bipyramid and aligned representation are used to recover the spatial information in order to achieve better performance. However, these operations make the network slower than two-stage methods such as Mask R-CNN. CenterMask \cite{lee2020centermask} decomposed the instance segmentation task into two parallel branches: Local Shape prediction branch, which is responsible for separating the instances, and Global Saliency branch to segment the image into a pixel-to-pixel manner. The branches are built upon a point representation layer containing the local shape information at the instance centers. The point representation is utilized from CenterNet \cite{zhou2019objects} for object detection with a DLA-34 as a backbone. Finally, the outputs of both branches are grouped together to form the final instance masks. CenterMask achieved mAP of 33.1\% on the COCO dataset at 25.2 fps using Titan Xp GPU. Although the one-stage methods run at a higher frame rate, the network speed is still an issue for real-time applications. The bottlenecks in the one-stage methods in the upsampling process in some methods, such as YOLACT.

\subsection{Bounding Polygon for Instance Segmentation}
\label{sec-poly-tech}

Recently, a new paradigm has been proposed to perform instance segmentation as predicting the polygon of each object instance through the center of the object. ExtremeNet \cite{zhou2019bottom} predicts an octagon mask for each instance that consists of 8 extreme points using a keypoint estimation network, achieving mAP of 34.6\% on COCO dataset. ExtremeNet 's backbone is Hourglass-104 \cite{newell2016stacked}, which is extremely heavy and requires a lot of time.
ESE-Seg \cite{xu2019explicit} estimated the shape of the detected objects by using an explicit shape encoding and decoding framework. It is based on Inner-center Radius (IR) and fits it using Chebyshev polynomial fitting, and YOLOv3 is the object detector. ESE-Seg archives mAP of 21.6\% on the COCO dataset.
PolarMask \cite{xie2020polarmask} was published concurrently with ESE-Seg but conducted independently. PolarMask proposed an additional polar IoU loss. Also, it does not require box detection, but it is necessary for ESE-Seg. PolarMask outperforms ESE-Seg by mAP of 7.5\% on the COCO dataset.

FourierNet \cite{benbarka2020fouriernet} uses polygon representation to represent each mask. It is a fully convolutional method with no anchor boxes, a shape vector is predicted and then converted into contour points using Fourier transform. The predicted boundaries are smoother than PolarMask. FourierNet achieves 24.3\% mAP on the COCO dataset at 26.6 FPS using 2080Ti GPU, which is not suitable for real-time applications. PolyTransform \cite{liang2020polytransform} combined the power of the current segmentation approaches and polygon-based methods to produce geometry-preserving masks. Instance masks are generated using segmentation network, and then the masks are converted to a set of polygons using deforming network to fit the object boundaries better. It achieved mAP of 44.6\% on the Cityscapes dataset when the model was trained on the COCO dataset. Our method Insta-YOLO follows this paradigm without any upsampling technique, unlike others. The polygon of each instance regresses using N point, where N is a hyperparameter representing the number of points for each instance, and we achieve comparable mAP at much higher FPS than other methods.


\section{Methodology}
\label{sec-methodology}
In this section, our proposed model is detailed. Data generation is demonstrated, followed by the network architecture and our loss functions.

\subsection{Data Generation}
\label{sec-data-generation}
 
To be able to train a conventional object detection algorithm, ground truth needs to be represented by the center of the box, width, and height which are only sufficient for box regression and not a polygon. On the other hand, segmentation ground truth is usually provided as dense pixel-wise annotation. Although Insta-YOLO predicts a dense mask output, it takes only object contours as inputs without dense annotation. Therefore, a conversion tool is implemented to adapt the segmentation masks to polygon representation with an arbitrary number of coordinate points as a pre-processing step. For this task, the algorithm in \cite{teh1989detection} is adopted to detect the dominant points on a curved shape without the need for hand-crafted parameters.

To be able to train Insta-YOLO, a fixed number of vertices is needed to represent the object mask. To achieve that, two ways are presented. The first one is simpler but less accurate. It relies on sampling the desired number of vertices with a fixed step which is calculated according to 

\begin{equation}\label{fixed_step}
   Step =\frac{\text{\# of dominant points}}{\text{\# of desired vertices}}
\end{equation}

The second method is more complex but efficient enough to guarantee the most accurate representation for the object mask.
Using \cite{douglas1973algorithms}, we shrink the number of points needed to represent the object mask. It is based on the Hausdorff distance between the curves to get the most representative simplified curve. The simplified curve consists of a subset of the points that defined the original curve. We use a parametric version from the algorithm which needs to define an $\epsilon$ that is an indication for how accurate the new simplified curve should be, a wise selection of $\epsilon$ is required to get the correct output. To avoid setting it manually, we use binary search technique to search for the best $\epsilon$ by changing the $\gamma$ as shown in Equ. \ref{eq_eq2}, which determines the highest accuracy with the desired number of points iteratively.

\begin{equation}\label{eq_eq2}
   \epsilon = \gamma_{B\cdot S} \ast \text{Arc - Length}
\end{equation}

This technique enables us to represent the object mask with an adaptive step instead of a fixed-step. The adaptive step could be considered an object-aware step adapted to the object curvature to fit it as much as possible with the desired number of vertices. Fig.\ref{fig-fixed_adaptive_step} demonstrates the importance of our proposed adaptive step polygon. Curves lines are represented with more dense points than straight lines. Using fixed-step representation, the vehicle's front curved side is represented by 3 points only, i.e; points 1,2,3 in red, while the bottom of the car is represented by 4 points, i.e; points 6,7,8,9 although it is a straight line which can be represented only by two points. On the other hand, adaptive-step represents the vehicle's front side as 4 points providing the maximum accuracy for the curve, and represents the bottom as two points only, i.e; points 10,11. This mechanism helps the network learn attention on curved lines maximizing the accuracy while minimizing redundant points for easier curves which maximizes efficiency.

To obtain the optimal number of points which best represents objects of the dataset, an automatic method is adopted for such computation. We set an initial value for the number of points and compute IoU with the segmentation mask. We increase the number of points iteratively until we observe no significant change in IoU.

\begin{figure}
  \centering
  \includegraphics[width=0.4\textwidth]{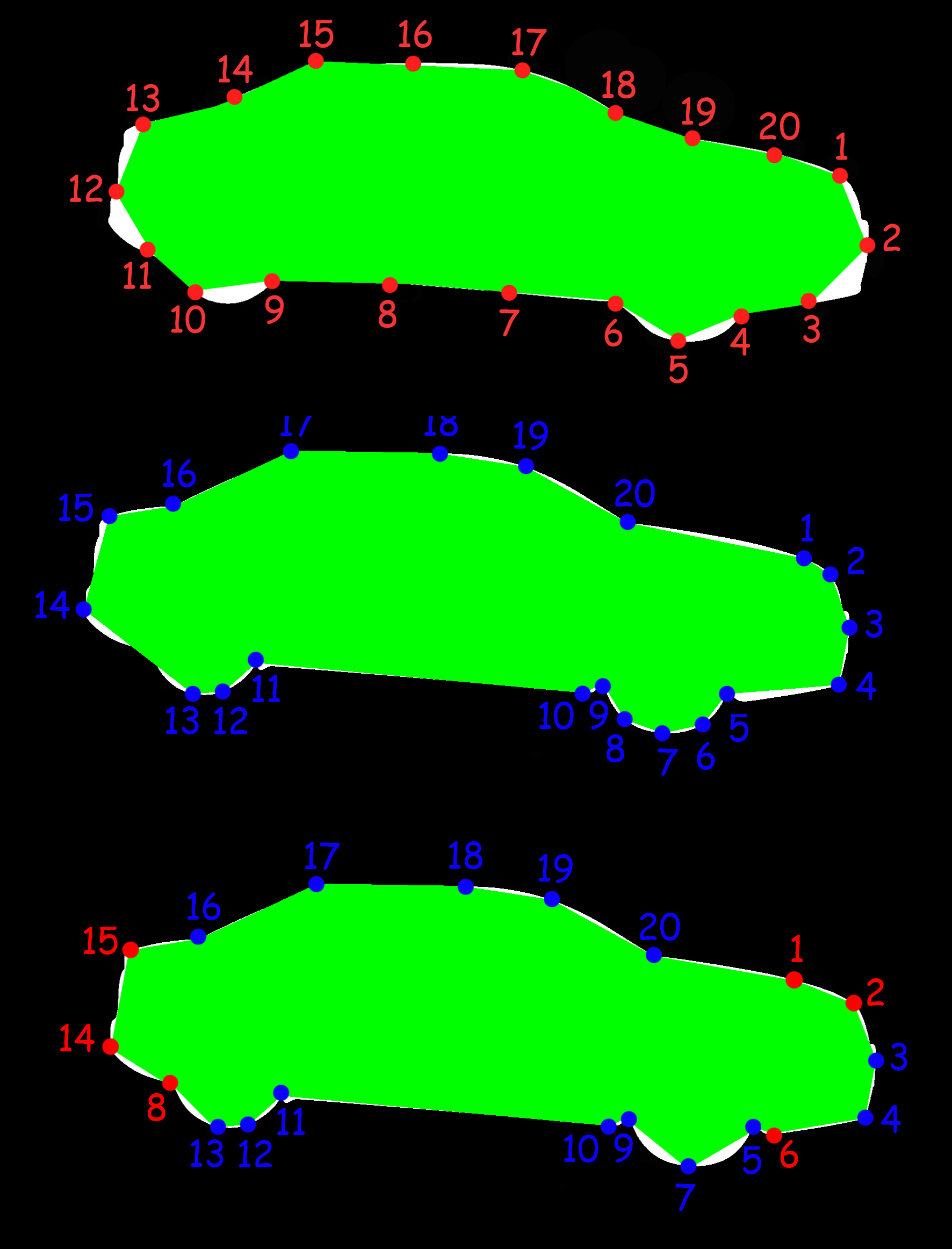}
  \caption{\textbf{Top:} Fixed representation for contour points. \textbf{Middle:} Variable step representation. Points 2,3 show the benefit of variable step compared to fixed steps. \textbf{Bottom:} Another valid representation demonstrating that there is no unique set of contour points for each instance}
  \label{fig-points}
\end{figure}


\subsection{Network Architecture}
\label{sec-Network-Architecture}

\begin{figure*}
  \centering
  \includegraphics[width=1\textwidth]{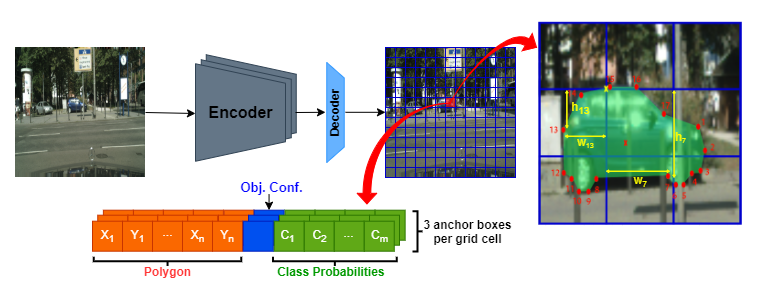}
  \caption{Insta-YOLO architecture which is inspired by YOLO, the right part illustrate our adaptation in the output layer.}
  \label{fig-network-arch}
\end{figure*}

Our backbone builds upon the popular YoloV3 \cite{redmon2018yolov3} algorithm as shown in Fig. \ref{fig-network-arch}. The output layer is adapted to follow equation \ref{eq-instayolo-outlayer} where the four coordinates were replaced by the number vertices which represent the object masks and the one is the mask confidence.

\begin{equation}
\label{eq-instayolo-outlayer}
output_layer=n_{anchros}[n_{vertices}1+n_{classes}]
\end{equation}

\subsection{Losses Representations}
\label{losses-Representations}

Our proposed method four loss functions as shown in Equ. \ref{eq_total-loss} that will be discussed in details in the following section.

\begin{equation}
\label{eq_total-loss}
\begin{split}
\text{Insta-YOLO loss} = \text{Classification loss} +\text{Confidence loss}\\ +\text{Localization loss}
\end{split}
\end{equation}

Two of which are inherited from YoloV3 \cite{redmon2016you} which are classification and confidence loss. We adapt the algorithm to our problem by adding two novel loss functions which are regression loss and IoU loss. We study the impact of our added loss functions and provide results in Tables \ref{Table:airbus-results},\ref{Table:carvana-results}, \ref{Table:cityscape-results}.

\subsubsection{Classification and Confidence loss}

If an object is detected, the classification loss at each cell is the squared error of the class conditional probabilities for each class
\begin{equation}\label{eq_eq4}
\text{Classification loss}=\sum_{i=0}^{S^2} \mathds{1}_i^{obj}\sum_{\text{c}\in \text{classes}} (p_i (c)- p_i^{'} (c))^2
\end{equation}
where \\
$\mathds{1}_i^{obj}=1$ if an object appears in cell $i$, otherwise 0.\\
$p_i^{'} (c)$ denotes the conditional class probability for class $c$ in cell $i$.

The confidence loss is measuring the abjectness of the box as follows

\begin{equation}
\label{eq_eq5}
\begin{split}
\text{Confidence loss}=\sum_{i=0}^{S^2}\sum_{j=0}^{B} \mathds{1}_{ij}^{obj}(C_i - C_i^{'})^2 \\
+ \lambda_{noobj}
\sum_{i=0}^{S^2}\sum_{j=0}^{B} \mathds{1}_{ij}^{noobj}(C_i - C_i^{'})^2
\end{split}
\end{equation}

where \\
$C_i^{'}$ is the box confidence score of the box $j$ in cell $i$.\\
$\mathds{1}_{ij}^{obj}=1$ if the $j$ th boundary box in cell $i$ is responsible for detecting the object, otherwise 0.\\
$\mathds{1}_{ij}^{noobj}$ is the complement of $\mathds{1}_{ij}^{obj}$.\\
$\lambda_{noobj}$ weights down the loss when detecting background.

\subsubsection{Localization Loss}

We modified the localization part, which is considered one of our contributions. The original YOLO \cite{redmon2018yolov3} applies mean square error as a regression loss for the center, width, and height of the box. In our case, we don't predict these measurements as we have polygons instead of boxes, as described in \ref{sec-Network-Architecture}. Accordingly, different localization losses are examined to fit our new problem.

\textbf{Regression Loss}

As opposed to conventional mean-square error function, using the Log Cosh loss which is approximately equal to $(x ^ 2) / 2$ for small $x$ and to $abs(x) - log(2)$ for large $x$. This means that it works mostly like the mean squared error, but will not be aggressively affected by outliers. We maintain the same advantages of Huber loss \cite {huber1973robust}, while keeping our function differentiable.

\textbf{IoU Loss}

Figure \ref{fig-points} demonstrates that there is no unique representation for the object mask using a fixed number of vertices. Both right and left sides of the figure show almost identical mask for the vehicle. However there is slight difference between the points representing the mask as highlighted by the red points. Both masks are valid and both set of points are also valid. On the other hand, regression loss forces the network to predict the the contour points exactly as the annotation. This means that if the points were slightly shifted on the same contour, they will be penalized although it is a valid solution. This behavior makes the regression loss not enough for our problem as it will mislead the network and negatively affect the learned features. Therefore, two IoU losses have been added to our localization loss to be more suitable to this problem.

First, we follow the loss provided by \cite{xie2020polarmask} where the the vertices are converted from Cartesian to polar. The approach relies on approximating the area that is produced by the predicted polygon and the ground truth polygon to spherical shape, which enables us to convert the exact equation \begin{equation}\label{eq_eq6}
\text{IoU}= \frac{\int_{0}^{2\pi}\frac{1}{2}\text{min}(d,d^\ast)^2 d\theta}{\int_{0}^{2\pi}\frac{1}{2}\text{max}(d,d^\ast)^2 d\theta}
\end{equation}

to a discretized and approximated version of it 
\begin{equation}\label{eq_eq7}
\text{Polar IoU Loss}=\log \frac{\sum_{i=1}^{n}d_{max}}{\sum_{i=1}^{n}d_{min}}
\end{equation}

Second, we introduce a new loss function as described in Equ.\ref{eq_area} to get the area of polygon and adapt it to represent Cartesian IOU loss as follows:
\begin{enumerate} 
\item Convert Cartesian vertices to polar representation by calculating the distance and theta for every vertex relative to the polygon center according to Equ.\ref{eq_distance} and Equ. \ref{eq-theta-equ}
\item Sort the Cartesian vertices clockwise or anticlockwise, in the ground truth and predicted polygon, in the same manner, w.r.t $\theta$.
\item Get the minimum and maximum vertices between the sorted vertices of GT and predicted ones w.r.t the calculated distance. The minimum values model the intersection between GT and predicted polygon and the maximum vertices represent the union.
\item The final IOU loss will be as in Equ.\ref{eq-car-iou-loss}
\end{enumerate}

\begin{strip}

\begin{equation}\label{eq_area}
\text{Area}=\left|\frac{(x_1 y_2 - y_1 x_2)+(x_2 y_3 - y_2 x_3)\cdots+(x_n y_1 - y_n x_1)}{2}\right|
\end{equation}

\begin{equation}\label{eq_distance}
\text{D}=\sqrt{(x_n-x_c)^2+(y_n-y_c)^2}
\end{equation}

\begin{equation}\label{eq-theta-equ}
\theta = \begin{cases} \mbox 180+tan^{-1}\left(\dfrac{y_n-y_c}{x_n-x_c}\right), & \mbox{if } x_n-x_c < 0 \hspace{0.25cm}\&\hspace{0.25cm} y_n-y_c > 0 \\
                      \mbox 180-tan^{-1}\left(\dfrac{y_n-y_c}{x_n-x_c}\right), & \mbox{if } x_n-x_c < 0 \hspace{0.25cm}\&\hspace{0.25cm} y_n-y_c < 0  \\
                      \mbox tan^{-1}\left(\dfrac{y_n-y_c}{x_n-x_c}\right), & \mbox{otherwise} 
                      \end{cases}
\end{equation}

\begin{equation}
\label{eq-car-iou-loss}
\text{Cartesian IoU Loss}=\log \left( \frac{(x_1^{max} y_2^{max} - y_1^{max} x_2^{max})+(x_2^{max} y_3^{max} - y_2^{max} x_3^{max})\cdots+(x_n^{max} y_1^{max} - y_n^{max} x_1^{max})}{(x_1^{min} y_2^{min} - y_1^{min} x_2^{min})+(x_2^{min} y_3^{min} - y_2^{min} x_3^{min})\cdots+(x_n^{min} y_1^{min} - y_n^{min} x_1^{min})}\right)
\end{equation}
\end{strip}

where\\
$x_n$ is the $x$ coordinates of vertex $n$ and $x_c$ is the $x$ coordinates of the center\\
$y_n$ is $y$ coordinate of the nth vertex and $y_c$ is the $y$ coordinates of the center\\
$N$ is the number of vertices.

We argue that our method is more accurate than \cite{xie2020polarmask}, which approximates an arbitrary polygon to a spherical shape to get its area, but our method doesn't make this aggressive approximation.
For fast convergence and to achieve the highest accuracy, we formulate the localization loss as a weighted combination of the regression loss and the IoU loss, as shown in Equ. \ref{eq_localizationloss}, at the beginning of the training the weight $\lambda$ is one and start decreasing while training to help the model focusing on outputting vertices near to the GT which is considered as warm-up epochs, after that start to inject the IoU loss gradually to make the model more sensitive and aware to the curvature information.
\begin{equation}\label{eq_localizationloss}
   \text{Localization loss} = \lambda \ast \text{Regression - Loss} + (1 - \lambda) \ast \text{IOU - Loss}
\end{equation}

IoU losses are known for their inaccurate predictions for self-intersecting polygons where one side crosses over another. To avoid this problem, we use the loss after several epochs where the model has learned already not to predict self-intersecting polygons.


\begin{table}[!ht]
\caption{Instance Segmentation Results on Cityscapes}
\centering
\label{Table:cityscape-results}
\begin{tabular}{cccc}
\hline
Experiment          & $AP_{50}$       & $AP_{75}$      & Frame rate  \\ \hline
Insta-YOLO with \\ regression loss    & 74.17\%     & 22\%    & \textbf{56 FPS}   \\ \hline

Insta-YOLO with \\ Polar IoU Loss   & 84\%  & 43\%    & \textbf{56 FPS}   \\ \hline

Insta-YOLO with \\ Cartesian IoU Loss   & 89\%  & 56\%    & \textbf{56 FPS}   \\ \hline

Polar-Mask   & 91.8\%  & 81.4\%    & 28 FPS    \\ \hline

YOLACT \\ \cite{bolya2019yolact}   & \textbf{93\%}  & \textbf{83\%}       & 32 FPS  \\ \hline

\end{tabular}
\end{table}

\section{Experimental Setup}
In this section, we provide details about the datasets we use. This is followed by description of the network setup to train our models.

\subsection{Datasets}
We evaluate our algorithm on three well-known datasets with different applications. Carvana \cite{carvana} and Cityscapes \cite{Cordts2016Cityscapes} focus on the application of vehicles detection and autonomous driving scenes. Each of both datasets consists of 5k images. Airbus Ship dataset \cite{airbus} focuses on aerial images of ships. We filter 13k images out of the whole dataset showing reasonably-sized ships. For all datasets, we split 80\% of the data for training and 20\% for validation.


\subsection{Training Setup}
In our experiments, YoloV3 \cite{redmon2018yolov3} model was used with pre-trained ResNet-50 encoder. Decoder has been adapted to predict multiple contour points. The models are trained end-to-end for 80 epochs with schedule weighted factor $\lambda$ according to \ref{eq:schedule_lamda}. This helps the model to focus on regression loss in beginning of training to avoid self-intersecting polygons. Later, IoU loss is used gradually to improve segmentation accuracy. Adam optimizer is used with learning rate of $1e^{-4}$. 

\begin{equation}\label{eq:schedule_lamda}
\lambda =\text{max}(0.7822 + \frac{0.3429}{epoch}, 0.2)
\end{equation}

\begin{table}[!ht]
\caption{Instance Segmentation Results on CARVANA}
\centering
\label{Table:carvana-results}
\begin{tabular}{cccc}
\hline
Experiment                        & $AP_{50}$ & $AP_{75}$ & Frame rate  \\ \hline
Insta-YOLO \\With regression loss   & 86\%  & 68\%    & \textbf{56 FPS}   \\ \hline

Insta-YOLO \\With Polar IoU Loss   & 92.73\%  & 82\%  & \textbf{56 FPS}  \\ \hline

Insta-YOLO \\With Cartesian IoU Loss   & \textbf{99\%}  & 84\%   & \textbf{56 FPS}  \\ \hline

YOLACT \\\cite{bolya2019yolact}   & \textbf{99\%}  & 98\%       & 32 FPS  \\ \hline

MaskRCNN   & \textbf{99\%}  & \textbf{98.5\%}       & 7 FPS  \\ \hline
\end{tabular}
\end{table}

\begin{figure}
  \centering
  \includegraphics[width=0.48\textwidth]{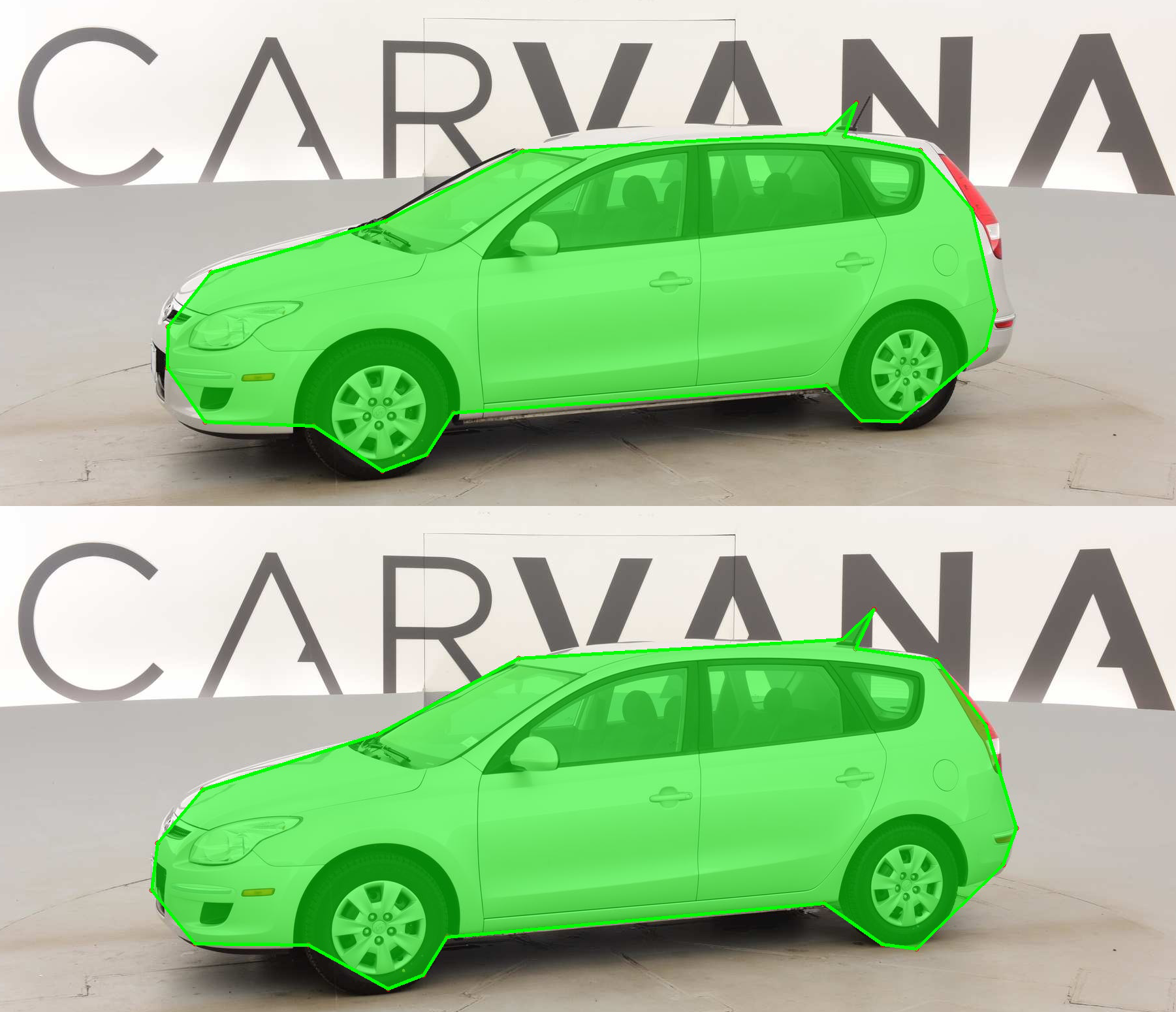}
  \caption{\textbf{Top:} Result on dataset using regression only. \textbf{Bottom:} Result using combined localization loss.}
  \label{fig-reg-vs-seg}
\end{figure}

\section{Experimental Results}
\label{sec-experiment}

In this section, we demonstrate qualitative and quantitative results of our approach for instance segmentation. This is followed by details about our experiments to predict oriented bounding boxes which is required for various applications.

\subsection{Instance Segmentation}

Tables \ref{Table:carvana-results} and \ref{Table:cityscape-results} illustrate the impact of our approach on Carvana \cite{carvana} and Cityscapes datasets \cite{Cordts2016Cityscapes}. With polar IoU from \cite{xie2020polarmask} loss we obtain improvement of 6\% and 10\% respectively in accuracy in both datasets relative to the baseline. Additional 7\% and 5\% improvement has been obtained using our proposed loss. Our model provides slightly less accuracy than Yolact, however it runs at 2.4 times the speed. For fair comparison with Yolact, the encoder of Yolact has been changed to DarkNet-53 which helps avoid misleading results. Figure\ref{fig-reg-vs-seg} visualizes the benefit of our proposed loss function which maximizes the accuracy of prediction.

\begin{figure}
  \centering
  \includegraphics[width=0.415\textwidth]{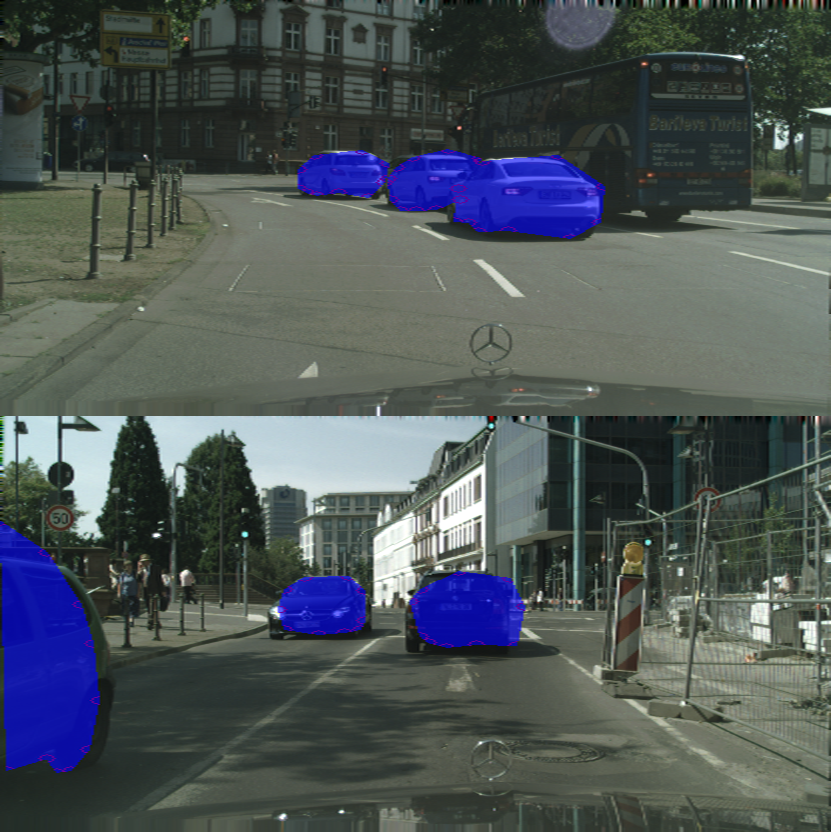}
  \caption{Visualization for Insta-YOLO output on Cityscapes dataset.}
  \label{fig-cityscapes}
\end{figure}

Fig.\ref{fig-reg-vs-seg} demonstrates the benefit of using our combined localization loss compared to regression only on Carvana dataset \cite{carvana}. Higher accuracy has been obtained using the combined loss. Fig.\ref{fig-cityscapes} shows our instance segmentation results on Cityscapes dataset. It is worth noting that our results are reported on "vehicles" class only.

\subsection{Oriented Bounding Boxes}

Various applications such as \cite{tang2017arbitrary} and \cite{ali2018yolo3d} require prediction of oriented bounding boxes. Such method requires an additional regression operation to be performed to predict the angle of the box. Conventional oriented boxes methods suffer from angle encoding problem. For practical implementation constraints, box angle is predicted in the range of 0-180 degrees. This means that a slight change of a box oriented with 2 degrees can cause the prediction jump from 2 to 178 which might mislead the network and cause aggressive jumps in prediction.

Insta-YOLO does not have this limitation. Since the four box points are independent of each other, the resulting polygon will be oriented without additional parameter to learn and free of the angle encoding problem. Table \ref{Table:airbus-results} shows that our algorithm supersedes \cite{ali2018yolo3d} in accuracy by 5\% and runs at 2.7 times the speed, while it provides competitive results compared to Yolact 
at 1.57 times the speed.

\begin{table}[!ht]
\caption{Oriented Bounding Boxes Results on AirBus Ships}
\centering
\label{Table:airbus-results}
\begin{tabular}{cccc}
\hline
Experiment       & $AP_{50}$        & $AP_{75}$         & Frame rate  \\ \hline
Insta-YOLO   & \textbf{78.16\%}  & 45\%    & \textbf{56 FPS}   \\ \hline

Orinted YOLO, \\ YOLO3D \\ \cite{ali2018yolo3d}   & 73.4\%      & 21.98\%       & \textbf{56 FPS}  \\ \hline

YOLACT \\ \cite{bolya2019yolact}      & 74\%       & \textbf{48\%}       & 32 FPS  \\ \hline

\end{tabular}
\end{table}



\section{Conclusion}
\label{sec-conclusion}

In this paper, we proposed a novel algorithm for instance segmentation. The algorithm runs in real-time at 35 fps which is 2.5 times the speed of state-of-the-art algorithms with competitive results in terms of accuracy. A new loss function has been proposed to adapt YoloV3 algorithm to our problem. Ablation study has been performed demonstrating the benefit of our proposed loss. Our algorithm is evaluated on three well-known datasets, namely, Cityscapes \cite{Cordts2016Cityscapes}, Carvana \cite{carvana} and Airbus-Ship datasets \cite{airbus}. Both qualitative and quantitative results on the three datasets demonstrate the benefit of our algorithm in terms of accuracy, speed and efficiency. The proposed algorithm is highly generic where it can replace conventional boxes at higher accuracy as well oriented boxes for various applications as illustrated in our results compared to state-of-the-art methods.

\medskip
{\small
\bibliographystyle{icml2020}
\newpage
\bibliography{camera_ready.bib}
}


\end{document}